\pdfoutput=1
\documentclass[letterpaper]{article} 
\usepackage{aaai21}  
\usepackage{times}  
\usepackage{helvet} 
\usepackage{courier}  
\usepackage[hyphens]{url}  
\usepackage{graphicx} 
\usepackage{natbib}  
\usepackage{caption} 
\usepackage{multirow}
\usepackage{booktabs}
\usepackage{threeparttable}
\usepackage{enumerate}
\usepackage{epsfig}
\usepackage{subfigure}
\usepackage{amsmath}
\usepackage{amssymb}
\usepackage{autobreak}
\usepackage{cite}
\title{SMA-STN: Segmented Movement-Attending Spatiotemporal Network for Micro-Expression Recognition}
\author {

        Jiateng Liu \textsuperscript{\rm 1},
        Wenming Zheng \textsuperscript{\rm 1}, 
        Yuan Zong \textsuperscript{\rm 1}\\
}
\affiliations {
    \textsuperscript{\rm 1} School of Biological Science and Medical Engineering, Southeast University, Nanjing, China\\
    jiateng$\_$liu@seu.edu.cn, wenming$\_$zheng@seu.edu.cn, xhzongyuan@seu.edu.cn
}

\begin{document}
\maketitle

\begin{abstract}
  Correctly perceiving micro-expression is difficult since micro-expression is an involuntary, repressed, and subtle facial expression,  
  and efficiently revealing the subtle movement changes and capturing the significant segments in a micro-expression sequence is the key to micro-expression recognition (MER). 
  To handle the crucial issue, in this paper, we firstly propose a dynamic segmented sparse imaging module (DSSI) to compute dynamic images 
  as local-global spatiotemporal descriptors under a unique sampling protocol, which reveals the subtle movement changes visually in an efficient way. 
  Secondly, a segmented movement-attending spatiotemporal network  (SMA-STN) is proposed to further unveil imperceptible small movement changes, 
  which utilizes a spatiotemporal movement-attending module (STMA) to capture long-distance spatial relation for facial expression and weigh temporal segments. 
  Besides, a deviation enhancement loss (DE-Loss) is embedded in the SMA-STN to enhance the robustness of SMA-STN to subtle movement changes in feature level. 
  Extensive experiments on three widely used benchmarks, i.e., CASME II, SAMM, and SHIC, show that the proposed SMA-STN achieves better MER performance than other state-of-the-art methods, 
  which proves that the proposed method is effective to handle the challenging MER problem.
\end{abstract}

\section{Introduction}
\begin{figure}
  \includegraphics[width=0.5\textwidth]{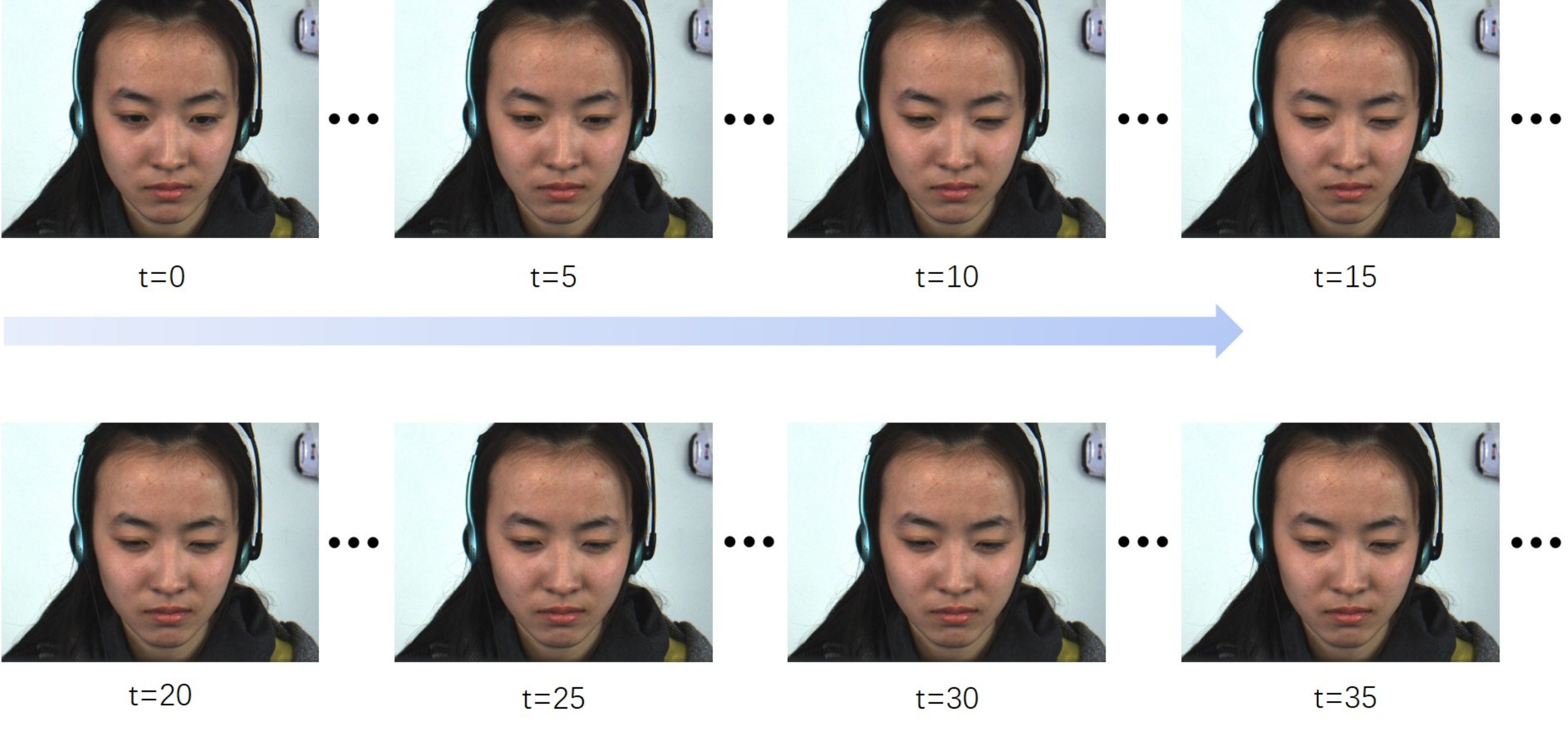}
  \caption{A sample sequence of micro-expression, where $t$ stands for time order. Since the subtle and involuntary changes of micro-expression, a number of frame samples are similar to each other,
  which are redundant for recognition.}
  \label{fig1}
\end{figure}
Micro-expression is a kind of unique facial expression usually occurring when people try to hide their genuine underlying emotions, which is subtle, involuntary, 
and has a short duration of only $\frac{1}{25}$s to $\frac{1}{5}$s \cite{ekman1969nonverbal}. Haggard et al. \cite{haggard1966micromomentary} first discovered the micro-expression during 
the research of ego mechanisms in psychotherapy.  And then, Ekman et al. \cite{ekman1969nonverbal} rediscovered this type of facial expression when looking at a video of psychopath and named 
it micro-expression officially. Unlike regular facial expressions, micro-expression is imperceptible and can reveal humans' real emotions what they want to conceal. Thus, automatically recognizing 
the micro-expression has many practical applications, such as criminal investigation, lie detection, and clinical diagnosis \cite{frank2009see}\cite{o2009police}\cite{frank2009behavior}. 

However, due to the characteristics of short duration and low intensity of micro-expression, realizing the automatic micro-expression recognition is still a very challenging task 
than conventional facial expression recognition problems, and even the person with professional training could get only $47\%$ accuracy in MER. \cite{frank2009see} As shown in Figure~\ref{fig1}, 
since the movement changes in a micro-expression sequence are incredibly subtle, many adjacent micro-expression data frames are similar to each other, 
and it is evident that capturing the segments which contains more subtle movement information can improve MER a lot.
According to \cite{TSN}, it is unnecessary to sample frames densely in a video sequence due to the information redundancy of adjacent frames. 
Hence, it is necessary to efficiently sample data to accurately extract spatiotemporal information of micro-expressions and suppress segments which contains less useful information for MER,  
and the traditional feature extraction methods for regular facial expressions are obviously not suitable to the micro-expressions. Furthermore, it is tough to discern the movement changes of micro-expressions  
for the the traditional spatiotemporal approaches due to the short duration. 
The aforementioned issues are usually ignored by the traditional handcraft methods or CNN-based approaches for MER.

To handle the crucial problems of MER, we first propose a novel dynamic representation extraction approach dubbed dynamic segmented sparse imaging module (DSSI) in this paper. 
Given a micro-expression sequence, DSSI firstly samples four sets of micro-expression data images from different segments of a sequence under a unique protocol. Each set contains three micro-expression 
images to compute a micro-expression dynamic images based on dynamic imaging method \cite{dynamicnetwork1}\cite{dynamicnetwork2}. DSSI can not only reveal subtle movement changes through capturing 
the micro-level spatiotemporal features of different segments in a micro-expression sequence, but also avoid the interference of useless data frames on MER. Secondly, we propose a segmented movement-attending 
spatiotemporal network (SMA-STN) to weigh different ME dynamic images and further unveil the subtle movement changes. The main components of SMA-STN include a CNN backbone and a spatiotemporal movement-attending module 
(STMA). The CNN backbone extracts spatiotemporal features from the computed dynamic images. And then, the spatiotemporal movement-attending module learns attention weights for different segments to capture both the 
global context information of a micro-expression sequence and the long-range dependencies of facial expression simultaneously by utilizing two types of self-attention mechanisms, including a global self-attention 
module and a non-local self-attention block. Subsequently, the processed dynamic features are aggregated to a global representation with calculated attention weights for the final classification. Furthermore, 
we also propose a deviation enhancement loss (DE-Loss) function embedded in the SMA-STN to further magnify the discrepancies of dynamic images generated from different segments of a micro-expression sequence, 
so that we can further improve the robustness of SMA-STN for subtle movement changes. Extensive experiments on three widely used micro-expression benchmarks show that the proposed model has strong robustness to 
subtle movement changes in a micro-expression sequence and can capture the representative spatial-temporal micro-expression features samples efficiently.

In a word, our contributions can be summarized as follows:
\begin{enumerate}[1)]
\item 
We propose a novel dynamic representation extraction approach termed dynamic segmented sparse imaging module (DSSI) to capture the subtle movement changes while reducing the redundant data for MER.
\item 
We propose a novel segmented movement-attending spatiotemporal network (SMA-STN) combined with a CNN backbone and a novel spatiotemporal movement-attending module (STMA) to extract representative features. 
SMA-STN can handle the crucial issue of capturing long-distance spatial relations of facial expression and weighting for different micro-expression sequence segments simultaneously.
\item 
A deviation enhancement loss (DE-Loss) function embedded in the SMA-STN is also proposed to magnify the micro-level movement discrepancies between different segments of a ME sequence further.
\end{enumerate}

\section{Related Work}
\subsection{Micro-Expression Recognition}
Previous researches about MER usually focus on extracting robust spatiotemporal features. In \cite{pfister2011recognising}, Pfister et al. firstly investigated 
the micro-expression recognition problem by using local binary pattern from three orthogonal planes (LBP-TOP) to describe the spatiotemporal characteristics 
of micro-expression video clips. Through their experimental results, the LBP-TOP feature is proved to be effective for the MER issue. Besides, they also utilized 
a temporal interpolation model (TIM) to normalize the number of micro-expression frames into a fixed size. In \cite{polikovsky2009facial}, Polikonsky et al. proposed 
a 3D-gradients orientation histogram-based feature descriptor to investigate the MER problem. In addition,
Wang et al.\cite{RPCA} used robust principal component analysis (RPCA) to extract the background information, and then both of LBP-TOP and Local Spatiotemporal Directional 
Features (LSDF) are used for MER. In \cite{MDMO}, Liu et al. proposed a sparse main directional mean optical-flow (Sparse MDMO) feature as a novel distance metric to 
learn a practical dictionary from micro-expression data samples, which can be computed easily.

With the development of deep learning recently, many researchers put forward a series of deep convolutional neural networks (CNNs) 
based methods to conduct MER, which achieved impressive results. For example, Verma et al. \cite{verma2019learnet} proposed 
a Lateral Accretive Hybrid network (LearNet) to spot the involuntary changes in a micro-expression sequence and classify them. In \cite{xia2019spatiotemporal}, 
Xia et al. first considered the spatiotemporal deformations of micro-expression samples by utilizing deep model named spatiotemporal recurrent convolutional 
networks (STRCN) with two types of extension. Furthermore, Song et al. \cite{ThreeStream} proposed a three-stream convolution network (TSCNN) combined with a dynamic-temporal stream,
a static-spatial stream and a local-spatial stream to aggregate temporal information and local region cues of micro-expressions for recognizing. 
Nevertheless, the inability to capture subtle movement changes and capture the signficiant segments of a micro-expression sequence in a efficient way is often neglected by these methods. 
\subsection{Attention Mechanisms In Deep Learning}
The attention mechanism is widely used in various fields including computer vision, natural language processing and so on \cite{Attention}. With the development of
deep learning in recent years, numerous works have applied attention mechanism to deep neural networks, which achieved excellent performance \cite{FAN}\cite{NAN}\cite{regionATT}.
In \cite{NAN}, Yang et al. proposed a Neural Aggregation Network (NAN) for video face recognition. They utilized attention to aggregate facial features 
of a person into a compact and fixed-dimension feature representation for recognition based on a deep convolutional neural network. 
Similarly, 
Wang et al.\cite{regionATT} proposed a region attention network (RAN) to handle the pose and occlusion variant facial expression recognition problem. They aggregated and 
embedded a varied number of region features produced by a self-attention module and a relation-attention module 
into a compact fixed-dimension feature representation before classification.
In \cite{nonlocal}, Wang et al. proposed several non-local attention frameworks to capture the long-range interactions among distant pixels in both spatial and temporal fields. 
The core idea of the non-local operation is also a self-attention mechanism and it achieved promising performance in many tasks, including video classification, 
object detection and pose estimation. 
Inspired by the various self-attention mechanisms, we design a novel spatiotemporal movement-attending module 
to capture the long-distance spatial relation for facial regions and learn importance weights for different segments of a micro-expression sequence at the same time.
\begin{figure}
  \center
  \includegraphics[width=0.45\textwidth]{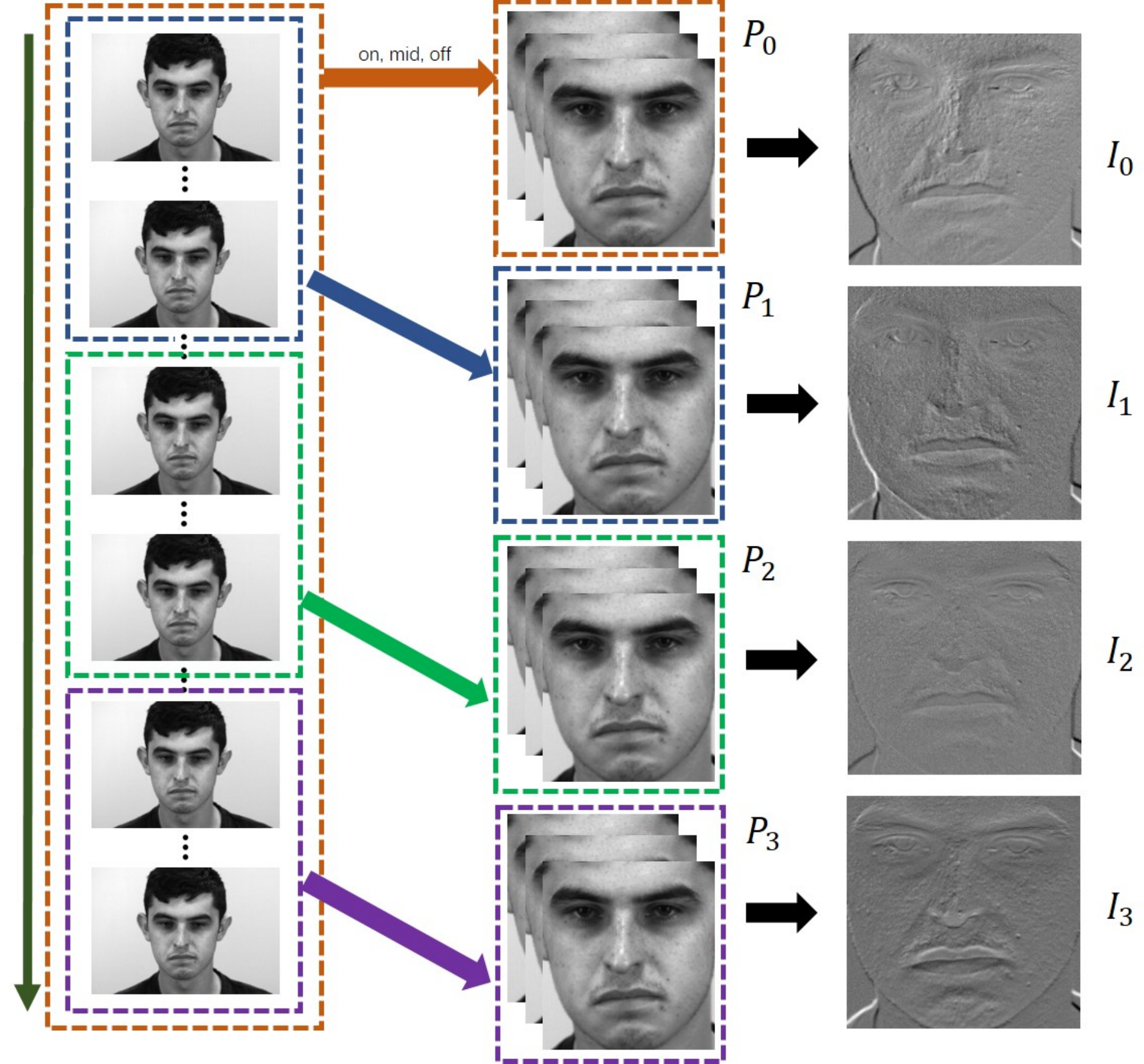}
  \caption{The illustration of DSSI. Given a sequence of micro-expressions, we first do the pre-processing for micro-expression samples. 
  Subsequently, DSSI divides the sequence into three segments evenly and sparsely samples three frames from each segment to generate three snippets $\{P_1, P_2, P_3\}$. Especially, a snippet $P_0$ consists of the onset, middle and offset frames of the sequence are
  also obtained. Finally, four dynamic images $\{I_0, I_1, I_2, I_3\}$ are computed by utilizing the obtained snippets in total.}
  \label{DSSI}
\end{figure}

\section{Proposed Methods}

\begin{figure*}
  \centering
  \includegraphics[height=0.33\textwidth,width=0.9\textwidth]{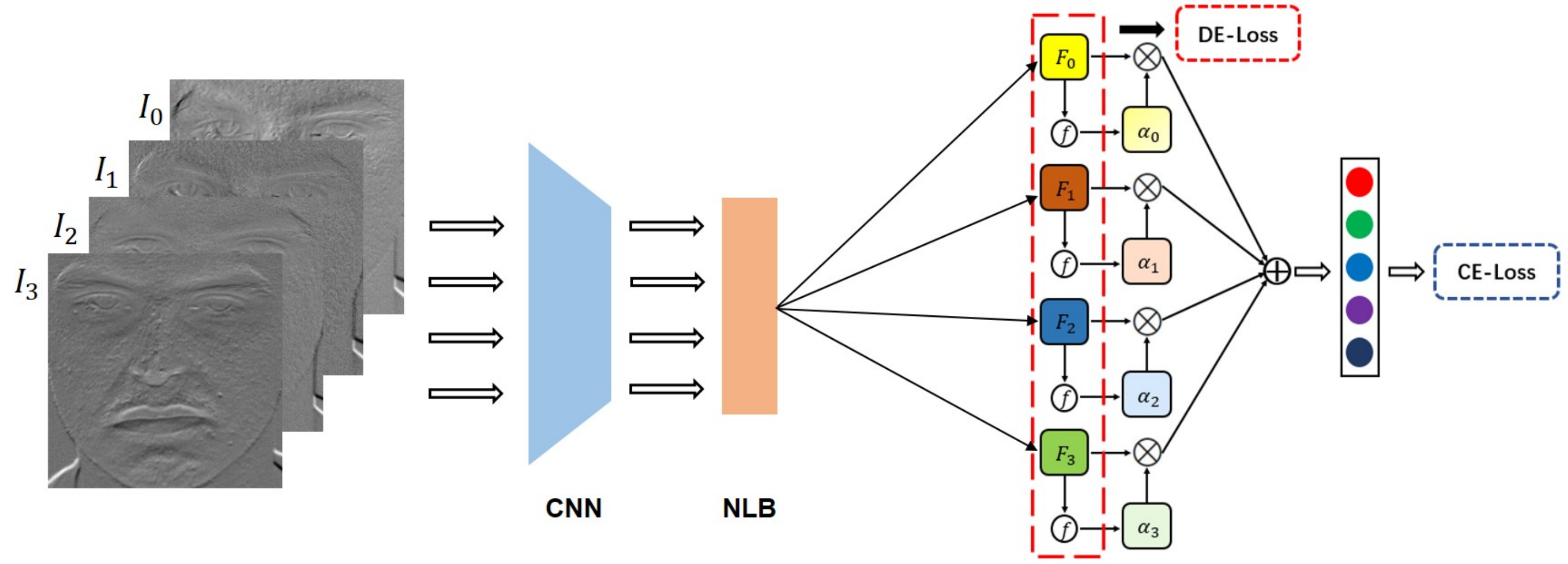}
  \caption{The pipeline of the proposed SMA-STN. Four dynamic images ($I_{0}, I_{1}, I_{2}, I_{3}$) calculated from a micro-expression sequence by DSSI are input 
  into a CNN backbone for feature extraction. Subsequently, the non-local block (NLB) calculates weights to capture long-range dependencies for facial regions 
  and further processes the features. And then, the global self-attention module learns weights for each dynamic image to obtain DE-Loss. A compact representation 
  is obtained after aggregating all the features with calculated attention ($\alpha_{0}, \alpha_{1}, \alpha_{2}, \alpha_{3}$) for final classification.}
  \label{stan}
\end{figure*}

This section describes the whole framework in detail.
As shown in Figure~\ref{DSSI}, given a sequence of micro-expression samples, DSSI computes four dynamic images as local-global spatiotemporal descriptors
under a special sampling protocol.
The computed micro-expression dynamic images are then fed into the CNN backbone for feature extraction as shown in Figure~\ref{stan}. Subsequently, the spatial non-local self-attention 
module processes the dynamic features to capture the long-range dependencies with a special non-local attention block. The global self-attention module weighs each segment by a fully-connected 
layer and the sigmoid function to enhance the most significant dynamic image. Besides, a deviation enhancement loss is further proposed to measure the micro-level discrepancies between different 
segments and magnify them. The final global representation is aggregated with the segment feature for classification.\\

\subsection{Dynamic Segmented Sparse Imaging}

According to \cite{TSN}, it would be efficient to sample frames with a sparse and global temporal strategy to capture the long-range video representation since consecutive frames are highly redundant.
Besides, the short duration of micro-expressions make a consequence that not all the micro-expression frames are useful to the micro-expression recognition, and the traditional 
feature extraction approaches for MER can not reflect overall information objectively. Motivated by the dynamic imaging 
technology \cite{verma2019learnet}\cite{dynamicnetwork1}\cite{dynamicnetwork2} and sparse sampling mechanism \cite{TSN}, we propose a novel dynamic representation extraction approach, namely dynamic 
segment sparse imaging (DSSI) module, to compute dynamic images in different segments as local-global spatiotemporal descriptors
so that we can capture the subtle movement changes of a micro-expression sequence while reducing redundant micro-expression frames. \\
\textbf{Sparse Sampling.} 
The illustration of dynamic segmented sparse sampling is shown in Figure~\ref{DSSI}. Given a sequence of micro-expression sample $V$, we first divide it into 3 segments $ \{S_1, S_2, S_3\} $ of equal 
duration without overlap after pre-process. For each segment, we then divide it into 3 equal sub-segment again, i.e. $S_i = \{s^1_i, s^2_i, s^3_i\}$.  
Finally, DSSI randomly samples a micro-expression instance snippet $P_i = \{P_{i}^{0}, P_{i}^{1}, P_{i}^{2}\}$ from each sub-segment in each segment. 
To capture global spatiotemporal information of a micro-expression sequence, a snippet including 
the onset, middle and offset frames of the sequence is also obtained as a comparison, i.e., $P_0 = \{P_{0}^{on}, P_{0}^{mid}, P_{0}^{off}\}$. \\
\textbf{Dynamic Imageing.} 
The core idea of dynamic imaging method is to represent a video by a single RGB image based on rank pooling \cite{dynamicnetwork1}\cite{dynamicnetwork2}. 
Supposing that a micro-expression sequence is represented as a ranking function by its frames $\{R_1, R_2, \dots, R_t\}$. Let $V_{i}=\frac{1}{i} \sum_{t=1}^{i} \psi\left(R_{t}\right)$ be time 
average of these features up to time t, where $\psi_t$ is the representation vector extracted from  each individual frame $R_t$ of the video sequence. 
Then the ranking function calculates a score to associate each time $t$ by $S(t|d)=<d,V_t>$, where $\mathbf{d} \in \mathbb{R}^{d}$. By learning the parameters of $d$, 
the scores can reflect the rank of the frames, i.e. $S(q|d)>s(t|d)$ when $q>t$. In the last, learning the $d$ can be considered as a convex optimization problem which can be solved 
by the RankSVM formulation:\\
$$
    \mathrm{d}^{*}=\rho\left(R_{1}, R_{2}, \ldots, R_{T} ; \psi\right)=\underset{\mathrm{d}}{\operatorname{argmin}} E(\mathrm{d}),
$$
$$
    E(\mathrm{d})=\frac{\lambda}{2}\|\mathrm{d}\|^{2}+ \\
     \frac{2}{T(T-1)} \times \sum \max \{0,1-S(q \mid \mathrm{d})+S(t \mid \mathrm{d})\},
  \eqno{(1)}
$$

Since it is computational to obtain a dynamic image, an practical approximation to rank pooling called approximate rank pooling (ARP) has been proposed
\cite{dynamicnetwork1}\cite{dynamicnetwork2}. The ARP is derived by considering the first step of Eq. 1 in a gradient-based optimization starting with
$\mathbf{d}^{*}=\overrightarrow{0}$, and we can obtain the first approximated solution as:
$\mathbf{d}^{*}=\overrightarrow{0}-\left.\eta \nabla E(\mathbf{d})\right|_{\mathbf{d}=\overrightarrow{0}} \propto-\left.\nabla E(\mathbf{d})\right|_{\mathbf{d}=0}$
for any $\eta > 0$, where\\
$$
\begin{aligned}
  \nabla E(\overrightarrow{0}) &\left.\propto \sum_{q>t} \nabla \max \{0,1-S(q \mid \mathbf{d})+S(t \mid \mathbf{d})\}\right|_{\mathbf{d}=\overline{0}} \\
  &=\sum_{q>t} \nabla\left\langle\mathbf{d}, V_{t}-V_{q}\right\rangle=\sum_{q>t} V_{t}-V_{q},
\end{aligned}
\eqno{(2)}  
$$
then $d^*$ can be expanded as follows:
$$
\mathbf{d}^{*} \propto \sum_{q>t} V_{q}-V_{t}=\sum_{t=1}^{T} \beta_{t} V_{t},
\eqno{(3)}
$$
Through expanding above formulation, the coefficients $\beta_{t}$ can be writing as scalar:
$$
\beta_{t}=2t-T-1,
\eqno{(4)}
$$
As shown in \cite{dynamicnetwork1}\cite{dynamicnetwork2}, the feature vector $\psi(R_t)$ can be replaced by individual video frames $R_t$. Thus, the dynamic image computation 
reduces to accumulating the time average of video frames after pre-multiplying them by $\beta_t$. The obtained $d^*$ has cumulative information which can be used as a 
spatiotemporal descriptor of a video sequence. 
Specially, four dynamic images $\{I_{0}, I_{1}, I_{2}, I_{3}\}$ related to different segments are calculated from four sampled snippets. From Figure~\ref{DSI_EMO}, we can find 
that although it is hard to distinguish different micro-expression images in a continuous sequence since the subtle movement changes, the calculated four dynamic 
images are significantly different from each other. That is to say, the subtle movement changes have been magnified successfully by the DSSI module visually.

\subsection{Spatiotemporal Movement-Attending Module}
\begin{figure}
  \centering
  \includegraphics[height=0.35\textwidth,width=0.38\textwidth]{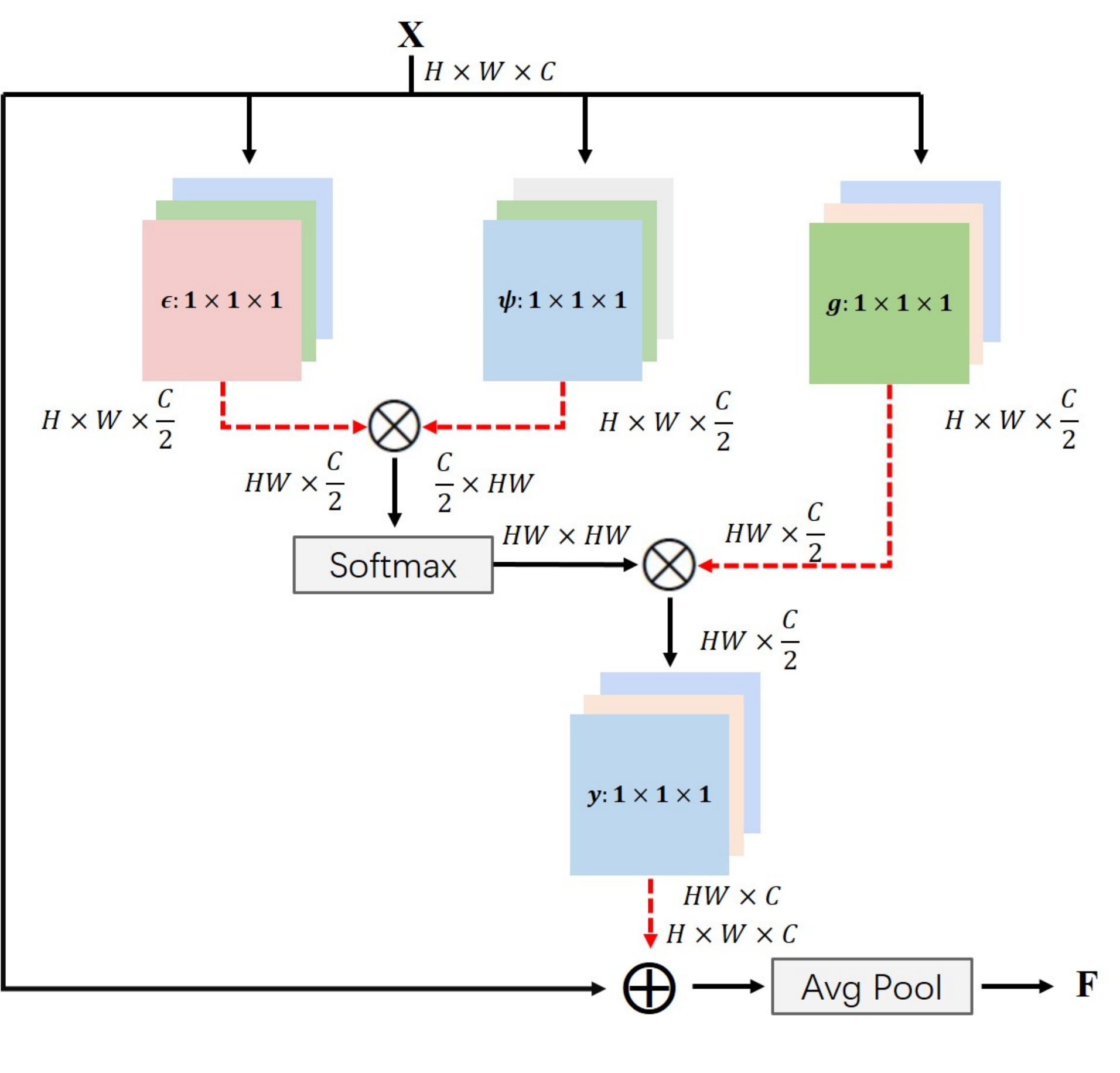}
  \caption{The illustration of the non-local block. \textcircled{$\times$} and \textcircled{+} denote matrix multiplication and elements-wise addition, respectively, and red dashed line
  denotes matrix reshaping.}
  \label{NLB}
\end{figure}

A spatiotemporal movement-attending module (STMA) set up with a spatial non-local self-attention block and a global self-attention module is proposed to capture long-distance spatial relation of facial regions 
and weight different segments of a micro-expression sequence at the same time.\\
\textbf{Spacial Non-local Self-Attention Block.} 
It is difficult to judge the type of expression based on only a portion of facial rerions according to the facial action units (AU) theory \cite{FACS}. 
 Thus, it is significant to capture the long-distance relation to consider 
 the comprehensive information of facial micro-expression to judge its emotion type objectively. To handle this problem,
 motivated by the 2D gaussian non-local networks \cite{nonlocal}, we utilize the embedded gaussian non-local self-attention 
 block added after the CNN backbone to capture the long-distance dependencies in extracted features. 
 It is obvious that the long-distance relation of micro-expression samples with same emotion category in a sequence are extremely similar to each other,
 hence the added non-local block shared weights with all the inputs, which is shown in Figure~\ref{NLB}.
 The non-local self-attention block is defined as follows:\\
$$
 f_{i}=W_{y} \frac{e^{\xi\left(X_{i}\right)^{T} \psi\left(X_{j}\right)}}{\sum_{j} e^{\xi\left(X_{i}\right)^{T} \psi\left(X_{j}\right)}} W_{g} X_{j}+X,
 \eqno{(5)}
$$
$$
 F_{i}=\rho(f_{i}),
 \eqno{(6)}
$$
where $\psi$ and $\xi$ are convolution operations with stride $1 \times 1$, $\rho$ denotes the average pooling operation, $W_y$ and $W_g$ are the weighting matrixes, and the input $X$ is the feature map 
 of a dynamic image extracted by the CNN. 
 The core idea of the non-local self-attention block is that more information can be maintained by 
 constructing a convolution operation whose size is the same as the feature map \cite{nonlocal}.\\
\textbf{Global Self-Attention Module.} 
It is believed that learning weights from both the local segments and global sequence is more beneficial to 
 classification \cite{FAN}. Thus, inspired by the frame-attention network \cite{FAN}, a segment-based global self-attention module is embedded after the spatial embedded gaussian 
 non-local self-attention block. To be specific, the weights of dynamic image related to the i-th segment can be calculated as:\\
$$
 \alpha_{i}=\operatorname{\emph{s}}\left(\operatorname{\sigma}\left(F_{i}^\top * q\right)\right),
 \eqno{(7)}
$$
  where $q$ denotes the parameters of a fully-connected layer, $\sigma$ denotes the sigmoid function, and \emph{s} denotes the softmax function. 
 All the input features are aggregated into a global feature representation with the attention weights as follows:\\
$$
 F_{m}= \sum_{i=0}^{n} \alpha_{i} F_{i},
 \eqno{(8)}
$$
$F_m$ is used as the representation for the final classification. Through the spatiotemporal movement-attending mechanism, not only the long-distance spatial relation of facial regions 
but the temporal movement changes can be considered together.\\

\subsection{Deviation Enhancement Regularization}
 To distinguish dynamic images related to different segments in feature level, 
 we propose a simple yet effective deviation enhancement loss function (DE-Loss) which magnifies the subtle movement changes among dynamic images by enhancing the standard deviation.
 We firstly measure differences between the four features extracted by the non-local sell-attention block
 through calculating and normalizing the Euclidean distance among them:
$$
 D_{i j}=\left\|F_{i}-F_{j}\right\|_{2},
 \eqno{(9)} 
$$
$$
 \overline{D_{i, j}}=\frac{D_{i, j}-D_{\operatorname{mean}}}{D_{\max }-D_{\min }},
 \eqno{(10)}
$$
 where $F_i$ and $F_j$ are the features of dynamic images extracted by the non-local sell-attention block ($i<j$).
 $D_{\max}$,  $D_{mean}$ and $D_{\min}$ denote the maximum value, the mean value and the minimum value of $D_{i, j}$, respectively. 
 The DE-loss is formulated by adding a margin to the standard deviation of $D_{i, j}$:
$$
 \mathcal{L}_{D E}=1-D_{s t d},
 \eqno{(11)}
$$
 In the training phase, the DE-Loss is jointly trained with the cross-entropy loss. Mathematically, the whole training loss can be formulated as follows:\\
$$
 \mathcal{L} = \mathcal{L}_{CE}+\lambda \mathcal{L}_{DE},
 \eqno{(12)}
$$
where $\lambda$ is the trade-off parameter. 

\section{Experiments}
In this section, we conduct extensive experiments on three public micro-expression databases to validate the performance of the proposed SMA-STN. Especially, 
we first introduce the used micro-expression databases and implementation details and then compare it to other state-of-the-art MER methods to show 
the effectiveness of the proposed SMA-STN.\\
\subsection{Databases and Implementation Details}
\textbf{CASME II} 
The CASME II database \cite{CASMEII} is built by Yan et al. from the Institute of Psychology, Chinese Academy of Sciences, which recorded 247 micro-expression 
examples with action units (AUs) labeled from 27 subjects. These micro-expression examples are collected in a high temporal resolution of 100 fps categorised 
into five emotion classes, i.e., \emph{Happiness} (32 samples), \emph{Surprise} (25 samples), \emph{Disgust} (64 samples), \emph{Repression} (27 samples), 
and \emph{Others} (99 samples).\\
\textbf{SAMM} 
The SAMM database \cite{SAMM} is collected by Davison et al. from Manchester City University, which contains 159 micro-expression samples recorded from 29 subjects, 
and 8 micro-expression classes collected at 200 fps. Note that since the number of some micro-expression classes in the SAMM databases is too small, we only use the micro-expression classes 
whose number is larger than 10 in the experiment, namely \emph{Anger} (57 samples), \emph{Contempt} (12 samples), \emph{Happiness} (26 samples), \emph{Surprise} 
(15 samples), and \emph{Others} (26 samples).\\
\textbf{SMIC} 
The SMIC database \cite{SMIC} is set up by Li et al. from the University of Oulu, Finland. It consists of 164 samples recorded from 16 subjects in $3$ classes 
of emotions, namely \emph{Positive} (51 samples), \emph{Negative} (70 samples), \emph{Surprise} (43 samples), respectively. Different from the CASME II database, 
the samples of SMIC database are divided to three parts according to the recording equipment, including a high-speed camera (HS) of 100 fps, a normal visual 
camera (VIS) of 25 fps, and a near-infrared camera (NIR) of 25 fps. The SMIC-HS database consists of 164 micro-expression clips recorded from 16 subjects, 
while the SMIC-VIS and SMIC-VIR both have 71 micro-expression samples from 8 subjects. In this paper, we use the micro-expression samples recorded by a high-speed camera, 
namely the samples in SMIC-HS database.\\
\begin{figure}
  \center
  \includegraphics[height=0.3\textwidth,width=0.5\textwidth]{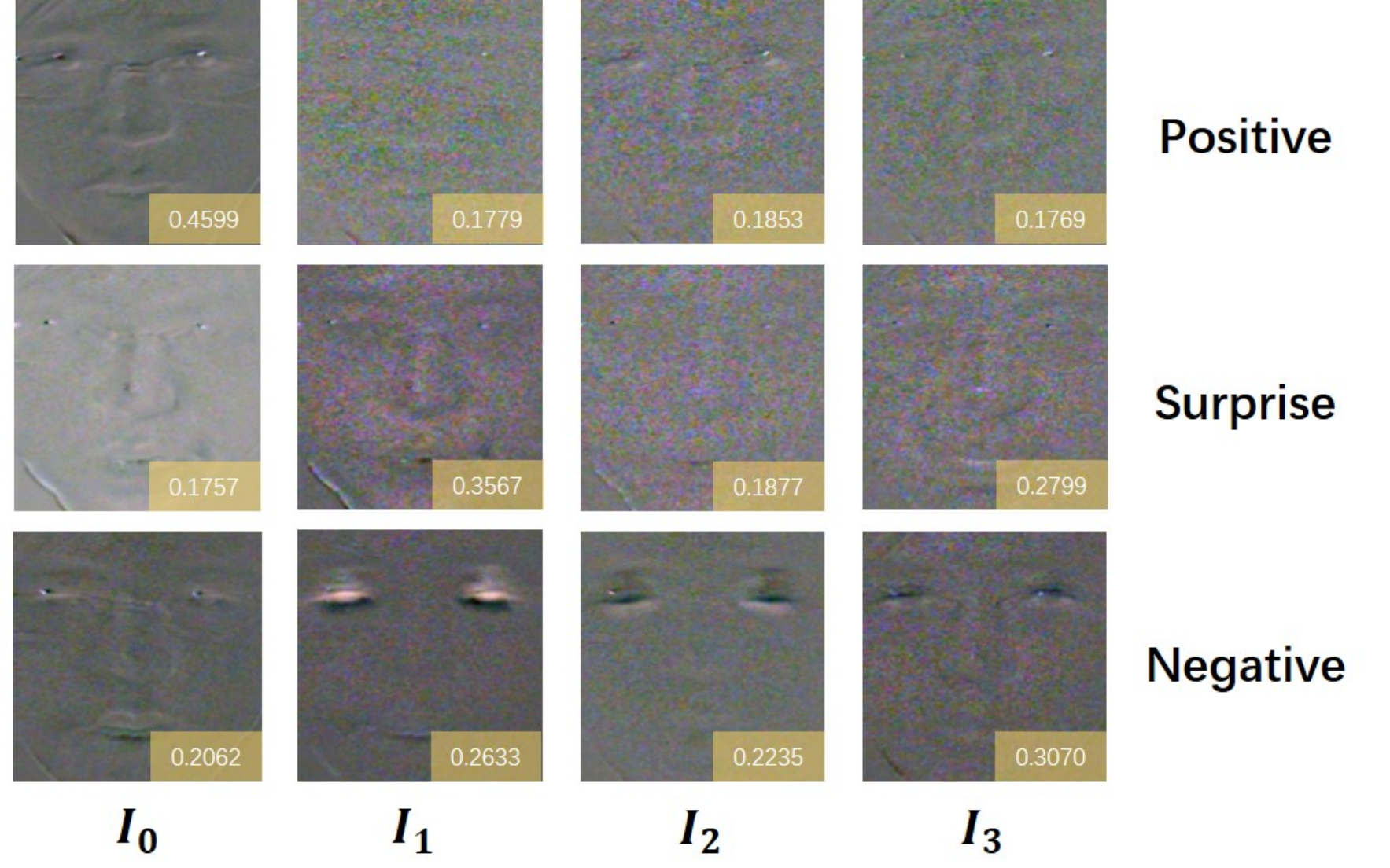}
  \caption{ Visualization of the learned $\alpha$ in SMA-STN. 
  Each row stands for a set of dynamic images computed by DSSI from a micro-expression sequence.   
  }
\label{DSI_EMO}
\end{figure}
\textbf{Implementation Details.}
In the experiments, the Face++ detection API is used to do the face detection and alignment, and then all the cropped 
micro-expression samples are resized to $224 \times 224$. The proposed SMA-STN is implemented with Pytorch toolbox, and we utilized 
the ResNet18 \cite{Resnethe2016deep} pre-trained on Face Attention Network \cite{faceattentionnetwork} as the CNN backbone. To avoid over fitting, the samples 
used for training are expanded with clockwise and counterclockwise rotation with $\pm 5$ and $\pm 10$ degree and horizontal flip.
In the training phase, the initial learning rate is set to $3 \times 10^{-4}$, and the training will be stopped in 100 epochs. The SMA-STN is trained in an 
end-to-end manner with Nvidia Titan Xp GPU. 
All the experiments are conducted under the leave-one-subject-out (LOSO) protocol since it can avoid samples 
from the same subject appearing in the training set and validation set, which can make the experimental results more reliable. Thus, only micro-expression 
data samples of a subject of the database are used for validating, and the remaining micro-expression data samples are used for training.\\
\begin{table}[htbp]
  \center
  \caption{The recognition accuracy and F1-score of different methods under the LOSO protocol on CASME II database.}
  \begin{threeparttable}
  \begin{tabular}{ccc}
  \hline
  Methods                     & Accuracy   & F1-score           \\ \hline
  LBP-TOP\cite{LBP-TOPCASME}  & 51.00 & 0.4700             \\
  EVM\cite{EVM+HIGO}          & 67.21 & N\textbackslash{}A \\
  STCLQP\cite{STCLQP}         & 58.39 & 0.5836             \\
  FDM\cite{FDM}               & 41.96 & 0.4700             \\
  SSSN\cite{DSSN}             & 71.19 & 0.7151             \\
  DSSN\cite{DSSN}             & 70.78 & 0.7297             \\
  TSCNN-I\cite{ThreeStream}   & 74.05 & 0.7327             \\
  TSCNN-II\cite{ThreeStream}  & 80.97 & 0.8070             \\
  \textbf{SMA-STN(Ours)} & \textbf{82.59} & 0.7946             \\ \hline
  \end{tabular}
  \begin{tablenotes}
    \footnotesize
    \item[*] N\textbackslash{}A - no results reported.
  \end{tablenotes}
\end{threeparttable}
\end{table}

\subsection{Evaluation of the SMA-STN}
\textbf{Comparison to the State-of-the-Art.} The experimental results of 3 databases under the LOSO protocol are shown in Table 1, Table 2 and Table 3, respectively.
Following the work in \cite{LearnFromHierzong2018learning} \cite{ThreeStream}, two evaluation methods including recognition accuracy and mean F1-score 
are adopted as the performance metric. 
To be specific, 
the F1-score is computed according to $F1-score=\frac{1}{c} \sum_{i=1}^{c} \frac{2 p_{i} \times r_{i}}{p_{i}+r_{i}}$,
where $p_i$ and $r_i$ stand for the precision and recall of the $i$th micro-expression, and $c$ denotes the number of micro-expression categories.\\
\begin{table}[htbp]
  \center
  \caption{The recognition accuracy and F1-score of different methods under the LOSO protocol on SAMM database.}
  \begin{threeparttable}
  \begin{tabular}{ccc}
  \hline
  Methods    & Accuracy   & F1-score \\ \hline
  LBP-TOP\cite{LBP-TOP-SAMM}    & 34.56      & 0.2892   \\
  LBP-SIP\cite{LBP-SIP}         & 36.03      & 0.3133   \\
  HOG-TOP\cite{EVM+HIGO}        & 36.03      & 0.3403   \\
  HIGO-TOP\cite{EVM+HIGO}       & 41.18      & 0.3920   \\
  SSSN\cite{DSSN}               & 56.62      & 0.4513   \\
  DSSN\cite{DSSN}               & 57.35      & 0.4644   \\
  TSCNN-I\cite{ThreeStream}     & 63.53      & 0.6065   \\
  TSCNN-II\cite{ThreeStream}    & 71.76      & 0.6942   \\
  \textbf{SMA-STN(Ours)} & \textbf{77.20} & \textbf{0.7033}   \\ \hline
  \end{tabular}
 \end{threeparttable}
\end{table}  
Besides, several state-of-the-art methods are used for comparison, including LBP-TOP \cite{LBP-TOPCASME, LBP-TOP-SAMM, LBP-TOP+TIM}, LBP-SIP \cite{LBP-SIP},
EVM + HIGO \cite{EVM+HIGO}, STCLQP \cite{STCLQP}, HIGO-TOP \cite{EVM+HIGO}, LPQ-TOP \cite{LPQ-TOP}, FMBH\cite{FMBH}, FDM \cite{FDM}, SSSN, DSSN \cite{DSSN} and TSCNN \cite{ThreeStream}. 
According to the results, we can find that the proposed SMA-STN achieves the best results in both recognition accuracy and F1-score in most cases. 
Specially, the proposed SMA-STN gets a better performance increase of 1.62$\%$ recognition accuracy compared to the state-of-the-art TSCNN-II method on CASME II database, 
and it achieves promising performance compared to the classic LBP-TOP with an increase of 31.59$\%$/0.3246 in recognition accuracy/F1-score.
As for SAMM database, the proposed SMA-STN obtains the best performance according to both recognition accuracy (77.20$\%$) and F1-score (0.7033).
Furthermore,  the proposed SMA-STN achieves the best results in both recognition accuracy (77.44\%) and F1-score (0.7683), which are higher than 
the state-of-the-art TSCNN-I method by a margin of 4.70\% and 0.0447 in recognition accuracy and F1-score on SMIC-HS database.\\
\textbf{Visualization of $\alpha$ in SMA-STN.} We visualize the learned importance $\alpha$ during training phase to investigate the effectiveness of SMA-STN to subtle movement changes. As shown in Figure~\ref{DSI_EMO},
we can find that the calculated dynamic images in a sequence are obviously different from each other, which indicates that the subtle movement changes in a micro-expression sequence are magnified.
Furthermore, SMA-STN efficiently enhances the weights of dynamic images which contains more movement information and suppresses the less ones.

\begin{table}[htbp]
  \center
  \caption{The recognition accuracy and F1-score of different methods under the LOSO protocol on SMIC database.}
  \begin{threeparttable}
  \begin{tabular}{ccc}
  \hline
  Methods                            & Accuracy   & F1-score           \\ \hline
  LBP-TOP + TIM\cite{LBP-TOP+TIM}    & 33.56      & N\textbackslash{}A \\
  STCLQP\cite{STCLQP}                & 64.02      & 0.6381             \\
  FDM\cite{FDM}                      & 54.88      & 0.5380             \\
  FMBH\cite{FMBH}                    & 71.95      & N\textbackslash{}A \\
  SSSN\cite{DSSN}                    & 63.41      & 0.6329             \\
  DSSN\cite{DSSN}                    & 63.41      & 0.6462             \\
  OFF-ApexNet\cite{OFF-ApexNet}      & 67.68      & 0.6709             \\
  TSCNN-I\cite{ThreeStream}          & 72.74      & 0.7236             \\
  \textbf{SMA-STN(Ours)}  & \textbf{77.44} & \textbf{0.7683}           \\ \hline
  \end{tabular}
  \begin{tablenotes}
    \footnotesize
    \item[*] N\textbackslash{}A - no results reported.
  \end{tablenotes}
 \end{threeparttable}
\end{table}
\begin{table}[htbp]
  \center
  \caption{Evaluation of the proposed attention and loss function modules in SMA-STN.}
  \begin{tabular}{cccccc}
  \hline
  Exp         & STMA     & EA-Loss      & Accuracy       & F1-score        \\ \hline
  1           & $\times    $  & $\times    $ & 68.90\%        & 0.6593          \\
  2           & $\times    $  & $\checkmark$ & 72.56\%        & 0.7218          \\
  3           & $\checkmark$  & $\times    $ & 74.39\%        & 0.7362          \\
  4           & $\checkmark$  & $\checkmark$ & \textbf{77.44\%} & \textbf{0.7683} \\ \hline
  \end{tabular}
\end{table}

\subsection{Ablation Study}
\begin{figure}
  \center
  \includegraphics[width=0.9\columnwidth]{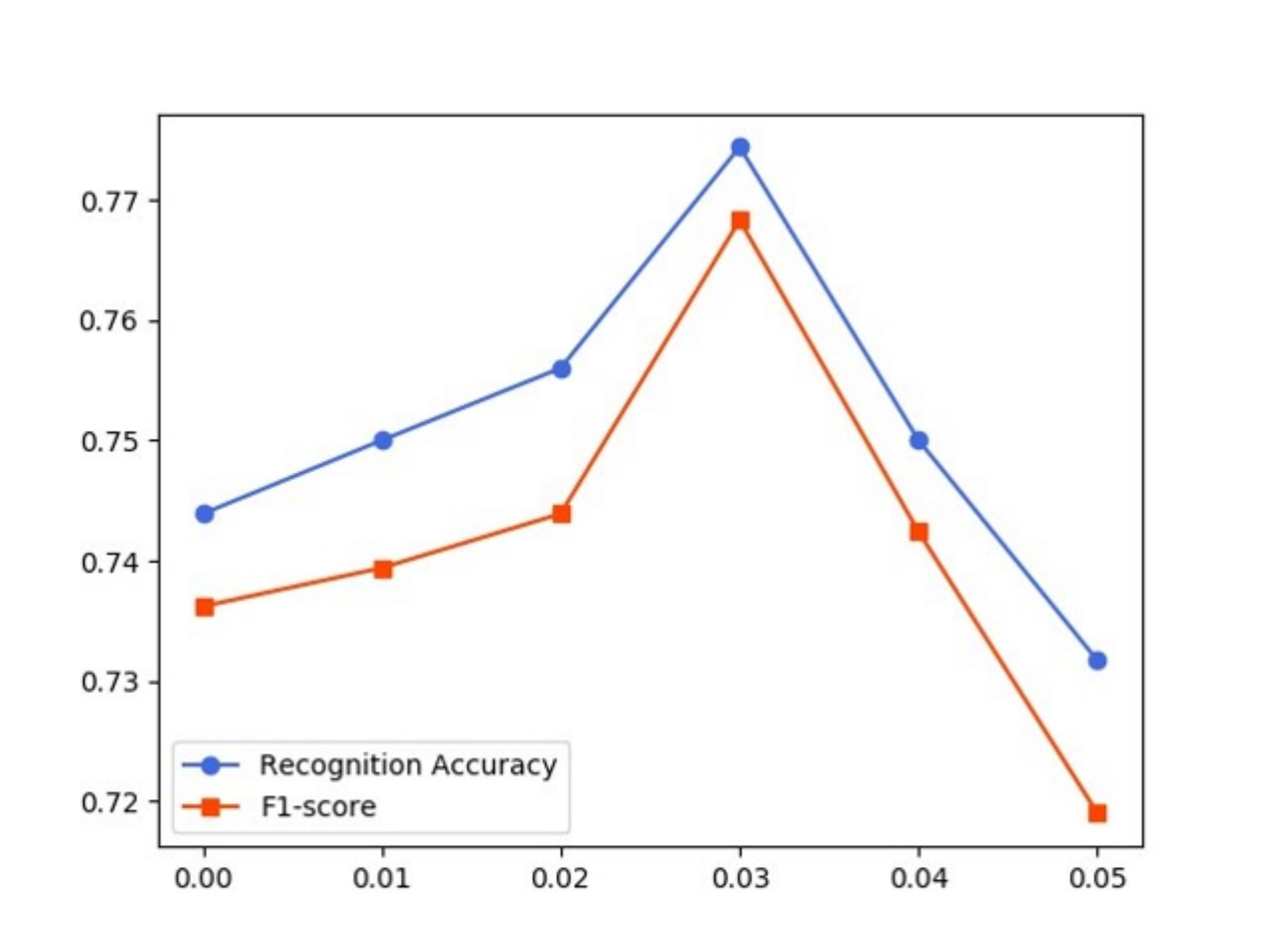}
  \caption{Evaluation of the tradeoff parameter $\lambda$ on the SMIC-HS database.}
  \label{linechart}
\end{figure}

An ablation study is conducted on the SMIC-HS database to investigate the generality of the proposed SMA-STN. \\
\textbf{Evaluation of the proposed modules in SMA-STN.} 
Several experiments are designed on the SMIC-HS database to evaluate the STMA and the DE-Loss in SMA-STN . 
The experimental results are presented in Table 4. According to the experimental results on recognition 
accuracy and F1-score, we can draw the following conclusions. First, all of the two modules improve the experimental results compared to the baseline (Exp.1) 
in varying degrees, which proves their effectiveness in improving MER issue performance. In addition, by adding three modules in turn, we achieve an improvement 
of 3.66\%/0.0625, 5.49\%/0.0769 and 8.54\%/0.109 in recognition and F1-score in Exp 2, 3 and 4 relative to the baseline experiments, which indicates that the 
combination of different modules can better improve the experimental results.\\
\textbf{Evaluation of the tradeoff parameter $\lambda$.} 
Several experiments are conducted to evaluate the effect of the difference value of $\lambda$ reported in Figure~\ref{linechart}. It shows that 
the most proper value of $\lambda$ to obtain a best result on SMIC-HS is 0.03, which obtains 77.44\% and 0.7683 in recognition accuracy and F1-score. Since $\lambda$ 
is a tradeoff coefficient to balance the classification cross-entropy and motion intensity magnification loss, the smaller values can not adequately distinguish 
subtle movement changes in different segments of a micro-expression sequence which results in lower experimental results. In contrast, a massive $\lambda$ value will make the network 
ignore the role of cross-entropy and lead to the degradation of classification performance.\\

\section{Conclusion}
In this paper, we propose an SMA-STN set up with a spatiotemporal movement-attending module (STMA) and a deviation enhancement loss (DE-Loss) to handle the crucial issue of revealing the subtle 
movement changes while enhancing the significant segments for MER in an efficient way. Besides, a novel dynamic segmented sampling imaging module (DSSI) is also proposed to compute dynamic images 
from micro-expressions sampled under a unique protocol, which can be considered as a spatiotemporal descriptor to capture the subtle movements while reducing the redundant data simultaneously. 
The STMA includes a spatial embedded gaussian non-local self-attention block and a global self-attention module to learn the long-distance spatial relation of facial regions and compute attention 
for divided segments and the whole sequence. The DE-Loss adds the standard deviation regularization in terms of Euclidean distance among extracted features to the SMA-STN, further improving the network's 
robustness to subtle movement changes in feature level. Through extensive experiments on three widely used micro-expression databases, the proposed SMA-STN outperforms other state-of-the-art methods, which proves 
its effectiveness on the MER issue.

\bibliography{STAN}
\end{document}